\pdfoutput=1
\documentclass[11pt]{article}

\usepackage[]{emnlp2021}

\usepackage{times}
\usepackage{latexsym}
\usepackage{enumitem}
\usepackage{microtype}
\usepackage{graphicx}
\usepackage{subfigure}
\usepackage{booktabs} %
\usepackage{times}
\usepackage{latexsym}
\usepackage{multirow}
\usepackage{color}
\usepackage{siunitx}
\usepackage{algorithmic, algorithm}
\usepackage{setspace}
\usepackage{multirow}
\usepackage{multicol}
\usepackage{titlesec}
\usepackage{boldline}
\usepackage{pifont}%
\usepackage{amsmath,amssymb,amstext, amsthm}
\definecolor{ForestGreen}{RGB}{34,139,34}
\graphicspath{{figures/}{../figures/}}

\usepackage{titlesec}

\titlespacing\section{0pt}{6pt plus 0pt minus 1pt}{2pt plus 1pt minus 1pt}
\titlespacing\subsection{0pt}{5pt plus 0pt minus 1pt}{0pt plus 0pt minus 0pt}
\titlespacing\subsubsection{0pt}{4pt plus 0pt minus 1pt}{1pt plus 0pt minus 0pt}
\titlespacing{\paragraph}{0pt}{4pt}{4pt}[3pt]

\usepackage{etoolbox}
\makeatletter
\preto{\@tabular}{\parskip=3pt}
\makeatother

\usepackage{caption}
 \newcommand{\fref}[1]{Figure~\ref{#1}}  
\newcommand{\tref}[1]{Table~\ref{#1}}

\def\e{{\mathbf e}} 
\def\x{{\mathbf x}} 
 
\def\y{{\mathbf y}}

\def\A{{\mathbf A}} 
\def\E{{\mathbf E}} 
 
\def\L{{\mathbf L}}

\newcommand{\method}[0]{\texttt{A2T} }
\newcommand{\methodV}[0]{\texttt{A2T-MLM} }
\newcommand{\base}[0]{vanilla adversarial training }

\newcommand{\TextAttack}[0]{\texttt{TextAttack} }
\newcommand{\fasttextfooler}[0]{\texttt{A2T} }
\newcommand{\fastbae}[0]{\texttt{A2T-MLM} }

\newcommand{\quotes}[1]{``#1''}
\DeclareMathOperator*{\argmax}{arg\,max}
\DeclareMathOperator*{\argmin}{arg\,min}
\newcommand{\importanceRankingName}{word importance ranking }

\usepackage[T1]{fontenc}

\usepackage[utf8]{inputenc}

\usepackage{microtype}

\title{Towards Improving Adversarial Training of NLP Models}

\author{Jin Yong Yoo, Yanjun Qi \\
  Department of Computer Science \\
  University of Virginia \\
  Charlottesville, VA, USA \\
  \texttt{\{jy2ma, yq2h\}@virginia.edu} \\}

\begin{document}
\maketitle

\begin{abstract}
Adversarial training, a method for learning robust deep neural networks, constructs adversarial examples during training. However, recent methods for generating NLP adversarial examples involve combinatorial search and expensive sentence encoders for constraining the generated instances. As a result, it remains challenging to use vanilla adversarial training to improve NLP models' performance, and the benefits are mainly uninvestigated. This paper proposes a simple and improved vanilla adversarial training process for NLP models, which we name Attacking to Training (\method\!). The core part of \method is a new and cheaper word substitution attack optimized for vanilla adversarial training. We use \method to train BERT and RoBERTa models on IMDB, Rotten Tomatoes, Yelp, and SNLI datasets. Our results empirically show that it is possible to train robust NLP models using a much cheaper adversary. We demonstrate that vanilla adversarial training with \method can improve an NLP model's robustness to the attack it was originally trained with and also defend the model against other types of word substitution attacks. Furthermore, we show that \method can improve NLP models' standard accuracy, cross-domain generalization, and interpretability. \footnote{Code is available at \url{https://github.com/QData/Textattack-A2T}}  
\end{abstract}

\section{Introduction}

Recently, robustness of neural networks against adversarial examples has been an active area of research in natural language processing with a plethora of new adversarial attacks\footnote{We use ``methods for adversarial example generation'' and ``adversarial attacks'' interchangeably.}  having been proposed to fool question answering \citep{jia2017adversarial}, machine translation \citep{Cheng18Seq2Sick}, and text classification systems \citep{Ebrahimi2017HotFlipWA, jia2017adversarial,alzantot2018generating, Jin2019TextFooler, pwws-ren-etal-2019-generating, pso-zang-etal-2020-word, garg2020bae}. One method to make models more resistant to such adversarial attacks is \textit{adversarial training} where the model is  trained on both original examples and  adversarial examples \cite{goodfellow2014explaining, madry2018towards}. Due to its simple workflow, it is a popular go-to method for improving adversarial robustness.

Typically, adversarial training involves generating adversarial example $\x'$ from each original example $\x$ before training the model on both $\x$ and $\x'$. In NLP, generating an adversarial example is typically framed as a combinatorial optimization problem solved using a heuristic search algorithm. Such an iterative search process is expensive. Depending on the choice of the search algorithm, it can take up to tens of thousands of forward passes of the underlying model to generate one example \citep{yoo2020searching}.  This high computational cost hinders the use of \base in NLP, and it is unclear how and as to what extent such training can improve an NLP model's performance ~\cite{Morris2020TextAttackAF}.

In this paper, we propose to improve the \base in NLP with a computationally cheaper adversary, referred to as \method\!. The proposed \method uses a cheaper gradient-based word importance ranking method to iteratively replace each word with synonyms generated from a counter-fitted word embedding \cite{Mrksic2016CounterfittingWV}. We use \method and its variation \methodV (which uses masked language model-based word replacements instead) to train BERT \citep{devlin2018BERT} and RoBERTa \citep{roberta} models on text classification tasks such as IMDB \citep{imdb_dataset}, Rotten Tomatoes \citep{pang2015MR}, Yelp \citep{Zhang2015Yelp}, and SNLI \citep{snli:emnlp2015} datasets. Our findings are as following:
\begin{itemize}[noitemsep,topsep=0pt]
    \item Adversarial training with both \fasttextfooler and \fastbae can help improve adversarial robustness, even against NLP attacks that were not used to train the model (see \tref{table:robustness-literature}).
    \item Adversarial training with \fasttextfooler can provide a regularization effect and improve the model's standard accuracy and/or cross-domain generalization, while \fastbae tends to hurt both standard accuracy and cross-domain generalization (see \tref{table:accuracy}).
    \item Using LIME \cite{lime} and AOPC metric, we demonstrate that adversarial training with \method can improve NLP models' interpretability (see ~\tref{table:aopc-scores}).
\end{itemize}

\section{Background}

\begin{figure}[]
\centering
\includegraphics[width=\hsize]{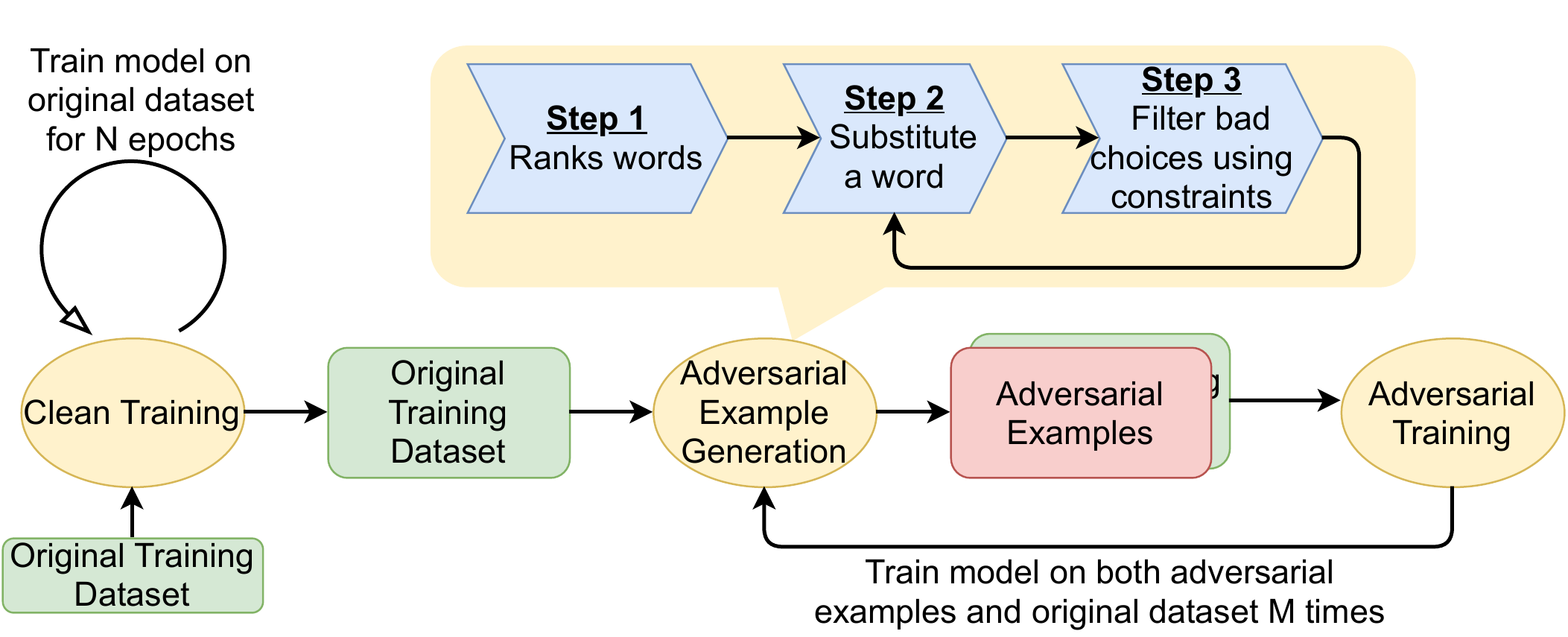}
\caption{Pipeline for vanilla adversarial training in NLP \label{fig:at}}
\end{figure}

\subsection{Vanilla Adversarial Training in NLP}

Vanilla adversarial training has been a major defense strategy in most existing work on adversarial robustness \citep{goodfellow2014explaining, kurakin, madry2018towards}. In our work, we define \textit{vanilla adversarial training} as adversarial training that involves augmenting the training data with adversarial examples generated from perturbing the training data in the input space. In contrast, non-vanilla adversarial training performs perturbations on non-input space such as word embeddings \cite{miyato2017adversarial, freeLB, jiang-etal-2020-smart, alum}.

In recent NLP literature, \base has only been evaluated in a limited context. In most studies, adversarial training is only performed to show that such training can make models more resistant to the attack it was originally trained with \citep{Jin2019TextFooler, pwws-ren-etal-2019-generating, li2020bertattack, pso-zang-etal-2020-word, clare2021}. This observation is hardly surprising, and it is generally recommended to use different attacks to evaluate the effectiveness of defenses \cite{carlini2019OnEvaluating}. Therefore, in this paper, we perform a more in-depth investigation into how a practical \base algorithm we propose affects NLP models' adversarial robustness against a set of different attacks that are not used for training.

In addition, we examine how adversarial training affects model's performance in other aspects such as standard accuracy, cross-domain generalization, and interpretability.

\subsection{Components of an NLP Attack}
\label{method:basics}
\fref{fig:at} includes a schematic diagram on \base where an adversarial attack is part of the training procedure. We borrow the framework introduced by \citet{morris2020reevaluating} which breaks down the process of generating natural language adversarial examples into three parts (see  \tref{table:faster-adv-attack}): (1) A search algorithm to iteratively search for the best perturbations (2) A transformation module to perturb a text input from $\x$ to $\x'$ (e.g. synonym substitutions) (3) Set of constraints that filters out undesirable $\x'$ to ensure that perturbed $\x'$ preserves the semantics and fluency of the original $\x$.
 
Adversarial attacks frame their approach as a combinatorial search because of the exponential nature of the search space. Consider the search space for an adversarial attack that replaces words with synonyms: If a given sequence of text consists of $N$ words, and each word has $M$ potential substitutions, the total number of perturbed inputs to consider is $(M + 1)^N - 1$. Thus, the graph of all potential adversarial examples for a given input is far too large for an exhaustive search. Studies on NLP attacks have explored various heuristic search algorithms, including beam search \cite{Ebrahimi2017HotFlipWA}, genetic algorithm \cite{alzantot2018generating}, and greedy method with \importanceRankingName \cite{ Gao2018BlackBoxGO,Jin2019TextFooler, pwws-ren-etal-2019-generating}.

\section{Method: \method (Attacking to Training)}
In this section, we present our algorithm \method for an improved and practical \base for NLP. We also present the cheaper adversarial attacks we propose to use in \method\!.

\subsection{Training Objective}

Following the recommendations by  \citet{goodfellow2014explaining, kurakin}, we use both clean\footnote{\textit{Clean} examples refer to the original training examples.} and adversarial examples to train our model. We aim to minimize both the loss on the original training dataset and the loss on the adversarial examples.

Let $\L(\theta, \x, \y)$ represent the loss function for input text $\x$ and label $\y$ and let  $\A(\theta, \x, \y)$ be the adversarial attack that produces adversarial example $\x_{adv}$. Then, our training objective is as following:
\begin{equation}
\begin{aligned}
    \argmin_{\theta} \E_{(\x,\y)\sim D}[\L(\theta, \x, \y) \\
    + \alpha \L(\theta, \A(\theta, \x, \y), \y)]
\end{aligned}
\end{equation}
$\alpha$ is used to weigh the adversarial loss. In this work, we set $\alpha=1$, weighing the two loss equally. \footnote{We leave tuning for the optimal $\alpha$ for future work.}

\subsection{A Practical  Training Workflow}
Previous works in adversarial training, especially those from computer vision \citep{goodfellow2014explaining, madry2018towards}, generate adversarial examples between every mini-batch and use them to train the model. However, it is difficult in practice to generate adversarial examples between every mini-batch update when using NLP adversarial attacks.

This is because NLP adversarial attacks typically require other neural networks as their sub-components (e.g. sentence encoders, masked language models). For example, \citet{Jin2019TextFooler} uses Universal Sentence Encoder \citep{Cer18USE} while \citet{garg2020bae} uses BERT masked language model \citep{devlin2018BERT}. Given that recent Transformers models such as BERT and RoBERTa models also require large amounts of GPU memory to store the computation graph during training, it is impossible to run adversarial attacks and train the model in the same GPU. We, therefore, propose to  instead maximize GPU utilization by first generating adversarial examples before every epoch and then using the generated samples to train the model.

Based on our initial empirical study, we found that it is not necessary to generate adversarial examples for every clean example in the training dataset to improve the robustness of the model. In fact, as we see in Section \ref{gamma-study}, training with fewer adversarial examples can often produce better results. Therefore, we propose to leave the number of adversarial examples produced in each epoch as a hyperparameter $\gamma$ where it is a percentage of the original training dataset. For cases where the adversarial attack fails to find an adversarial example, we skip them and instead sample more from the training dataset to compensate for the skipped samples. In our experiments, unless specified, we attack $20\%$ of the training dataset, which was based on our initial empirical findings.

Algorithm \ref{adv-training-algo} shows the proposed \method adversarial training algorithm in detail. We run clean training for  $N_{\text{clean}}$ number of epochs before performing $N_{\text{adv}}$ epochs of adversarial training. Between line 6-13, we generate the adversarial examples until we obtain $\gamma$ percentage of the training dataset. When multiple GPUs are available, we use data parallelism to speed up the generation process. We also shuffle the dataset before attacking to avoid attacking the same sample every epoch.

\begin{algorithm}[t]
\caption{Adversarial Training with \method}
\begin{algorithmic}[1]
    \REQUIRE Number of clean epochs $N_{\text{clean}}$, number of adversarial epochs $N_{\text{adv}}$, percentage of dataset to attack $\gamma$, attack $\A(\theta, \x, \y)$, and  training data $D=\{(\x^{(i)}, \y^{(i)})\}^n_{i=1}$, $\alpha$ the smoothing proportion of adversarial training 
    \STATE Initialize model $\theta$
    \FOR {clean epoch$=1,\dots, N_{\text{clean}}$}
        \STATE Train $\theta$ on $D$
    \ENDFOR
    \FOR {adversarial epoch$=1,\dots, N_{\text{adv}}$}
        \STATE Randomly shuffle $D$
        \STATE $D_{\text{adv}} \leftarrow \{\}$
        \STATE $i \leftarrow 1$
        \WHILE{$|D_{\text{adv}}| < \gamma * |D|$ and $i \leq |D|$}
            \STATE $\x^{(i)}_{\text{adv}} \leftarrow \A(\theta, \x^{(i)}, \y^{(i)})$
            \STATE $D_{\text{adv}} \leftarrow D_{\text{adv}} \cup \{(\x^{(i)}_{\text{adv}}, \y^{(i)})\}$
            \STATE $i \leftarrow i + 1$
        \ENDWHILE
        \STATE $D' \leftarrow D \cup D_{\text{adv}}$
        \STATE Train $\theta$ on $D'$ with $\alpha$ used to weigh the loss
    \ENDFOR
\end{algorithmic}
\label{adv-training-algo}
\end{algorithm}

\subsection{Cheaper Attack for Adversarial Training}
The attack component in \method is designed to be faster than previous attacks from literature. We achieve the speedup by making two key choices when constructing our attack: (1) Gradient-based word importance ordering, and (2) DistilBERT \citep{distil-bert} semantic textual similarity constraint. Table \ref{table:faster-adv-attack} summarizes the differences between \method and two other attacks from literature: TextFooler \cite{Jin2019TextFooler} and BAE \cite{garg2020bae}. %

\begin{table*}[t]
\centering
\scalebox{0.78}{
\begin{tabular}{ccccc}
\toprule[0.25ex]
Components & \fasttextfooler & \fastbae & TextFooler & BAE \\
\cmidrule(lr){1-5}
\begin{tabular}[c]{@{}c@{}}Search Method\\ for Ranking Words\end{tabular} & \begin{tabular}[c]{@{}c@{}}\textbf{Gradient-based}\\ Word Importance\end{tabular} & \begin{tabular}[c]{@{}c@{}}\textbf{Gradient-based} \\ Word Importance\end{tabular} & \begin{tabular}[c]{@{}c@{}}\textbf{Deletion-based}\\ Word Importance\end{tabular} & \begin{tabular}[c]{@{}c@{}}\textbf{Deletion-based}\\ Word Importance\end{tabular} \\
\cmidrule(lr){1-5}
Word Substitution & \textbf{Word Embedding }&  \textbf{BERT MLM } &  \textbf{Word Embedding} & \textbf{BERT MLM}  \\
\cmidrule(lr){1-5}
\multirow{2}{*}{ Constraints } & POS Consistency & POS Consistency & POS Consistency & POS Consistency \\
\cmidrule(lr){2-5}
& \textbf{DistilBERT Similarity}  & \textbf{DistilBERT Similarity} & \textbf{USE Similarity} & \textbf{USE Similarity}\\
\bottomrule[0.25ex]
\end{tabular}
}
\caption{Comparing \method and  variation with  TextFooler~\cite{Jin2019TextFooler} and BAE~\cite{garg2020bae}}
\label{table:faster-adv-attack}
\end{table*}

\begin{table}[]
\centering
\begin{tabular}{cr}
\toprule[0.25ex]
Attack & Runtime (sec) \\
\cmidrule(lr){1-2}
\fasttextfooler & 5,988 \\
TextFooler + Gradient Search  & 8,760 \\
TextFooler & 17,268 \\
\bottomrule[0.25ex]
\end{tabular} 
\caption{The runtime (in seconds) of \fasttextfooler\!, original TextFooler, and TextFooler where deletion-based word importance ranking is switched with gradient-based word importance ranking. We can see that replacing the search method gives us approximately $2\times$ speedup. The attack was carried out against a BERT model trained on IMDB dataset and 1000 samples were attacked.}
\label{tab:speedup-table}
\end{table}

\paragraph{Faster Search with Gradient-based Word Importance Ranking:}
Previous attacks such as \citet{Jin2019TextFooler, garg2020bae} iteratively replace one word at a time to generate adversarial examples. To determine the order of words in which to replace, both \citet{Jin2019TextFooler, garg2020bae} rank the words by how much the target model's confidence on the ground truth label changes when the word is deleted from the input. We will refer to this as deletion-based word importance ranking.

One issue with this method is that an additional forward pass of the model must be made for each word to calculate its importance. For longer text inputs, this can mean that we have to make up to hundreds of forward passes to generate one adversarial example.

\method instead determines each word's importance using the gradient of the loss. For an input text including $n$ words: $\x=(x_1, x_2, \dots, x_n)$ where each $x_i$ is a word, the importance of $x_i$ is calculated as:
\begin{equation}
    I(x_i) = ||\nabla_{\e_i} \L(\theta, \x, \y))||_1
\end{equation}
where $\e_i$ is the word embedding that corresponds to word $x_i$. For BERT and RoBERTa models where inputs are tokenized into sub-words, we calculate the importance of each word by taking the average of all sub-words constituting the word.

This requires only one forward and backward pass and saves us from having to make additional forward passes for each word. \citet{yoo2020searching} showed that the gradient-ordering method is the fastest search method and provides competitive attack success rate when compared to the deletion-based method. Table \ref{tab:speedup-table} shows that when we switch from deletion-based ranking (\quotes{TextFooler}) to gradient-based ranking (\quotes{TextFooler+Gradient Search}), we can obtain approximately $2\times$ speedup.

\paragraph{Cheaper Constraint Enforcing with DistilBERT \citep{distil-bert} semantic textual similarity model: } Most recent attacks like
\citet{Jin2019TextFooler, garg2020bae, li2020bertattack} use Universal Sentence Encoders (USE) \cite{Cer18USE} to compare the sentence encodings of original text $\x$ and perturbed text $\x'$. If the cosine similarity between  two encodings fall below a certain threshold, $\x'$ is ignored. 
One of the challenges of using large encoders like USE is that it can take up significant amount of GPU memory -- up to 9GB in case of USE. 

Instead of using USE, \method uses DistilBERT \citep{distil-bert} model trained on semantic textual similarity task as its constraint module \footnote{We use code from \citet{distilbert-sts}}. This is because DistilBERT requires $10\times$ less GPU memory than USE and requires fewer operations.

\subsection{\methodV: Variation with a Different Word Substitution Strategy}
\method generates replacements for each word by selecting top-$k$ nearest neighbors in a counter-fitted word embedding \cite{Mrksic2016CounterfittingWV}, which helps nearest-neighbor searches return better synonyms than regular word embeddings. This word substitution strategy has been previously proposed by \citet{alzantot2018generating, Jin2019TextFooler}. In our work, we first precompute all the top-$k$ nearest neighbors and cache them to speed up our attacks. We also consider another variation we name as \methodV in which BERT masked language model is used to generate replacements (proposed in \citet{garg2020bae, li2020bertattack, clare2021}).

We consider this variation because two strategies prioritize different language qualities when proposing word replacements. Counter-fitted word embeddings are likely to propose synonyms as replacements, but could produce incoherent texts as it does not take the entire context into account. On the other hand, BERT masked language model is more likely to propose replacement words that preserve grammatical and contextual coherency but fail to preserve the semantics. Comparing \method with \methodV allows us to study the effect of word substitution strategy on adversarial training.

\section{Related Work}
Past works on adversarial training for NLP models come in diverse flavors that differ in how adversarial examples are generated. \citet{miyato2017adversarial}, which is one of the first works to introduce adversarial training to NLP tasks, perform perturbations in the word embedding level instead of the actual input space level. Likewise, \citet{freeLB, jiang-etal-2020-smart, alum} all apply perturbations in the embedding level using gradient-based optimization methods from computer vision.

Another family of work on adversarial training involves computing the hyperspace of activations that contains all texts that can be generated using word substitutions and then training the model to make consistent prediction for inputs inside the hyperspace. \citet{jia-etal-2019-certified, huang-etal-2019-achieving} compute axis-aligned hyper-rectangles and leverages Interval Bound Propagation \cite{ibp} to defend the model against substitution attacks while \citet{dong2021towards} computes the desired hyperspace as a convex hull in the embedding space and further trains the model to be robust against worst case embedding in the convex hull.

Yet, adversarial training that simply uses adversarial examples generated in the input space is still a relatively unexplored area of research despite its simple, extendable workflow. Most works that have discussed such form of adversarial training only train limited number of models and datasets to show that adversarial training can make models more resistant to the particular attack used to train the model \citep{Jin2019TextFooler, pwws-ren-etal-2019-generating, li2020bertattack, pso-zang-etal-2020-word, clare2021}. Our work demonstrates that simple vanilla adversarial training can actually provide improvements in adversarial robustness across many different word substitution attacks. Furthermore, we show that it can improve both generalization and interpretability of models, properties that have not been examined by previous works.

\section{Experiment and Results}
\subsection{Datasets \& Models}
We chose IMDB \cite{imdb_dataset}, Movie Reviews (MR) \cite{pang2015MR}, Yelp \cite{Zhang2015Yelp}, and SNLI \cite{snli:emnlp2015} datasets for our experiment. For Yelp, instead of using the entire training set, we sampled 30k examples for training and 10k for validation.

\begin{table}[]
\centering
\begin{tabular}{lccc}
\toprule[0.25ex]
Dataset & Train & Dev & Test \\
\cmidrule(lr){1-4} 
IMDB & 20k   & 5k  & 25k  \\
MR   & 8.5k  & 1k  & 1k   \\
Yelp & 30k   & 10k & 38k  \\
SNLI & 550k  & 10k & 10k  \\
\bottomrule[0.25ex]
\end{tabular}
\caption{Overview of the datasets.}
\vspace{-5mm}
\end{table}

We trained BERT \cite{devlin2018BERT} and RoBERTa \cite{roberta} models using the implementation provided by \citet{wolf-etal-2020-transformers}. All texts were tokenized up to the first 512 tokens and we trained the model for one clean epoch and three adversarial epochs. Adam optimizer with weight decay of $0.01$ \cite{AdamW} and learning rate of $5\mathrm{e}{-5}$ were used for training. Also, we used a linear scheduler with $500$ warm-up steps for IMDB and Yelp, $100$ steps for MR, and $5000$  steps for SNLI. We performed three runs with random seeds for each model.

\subsection{Baselines}
Adversarial training can be viewed as a data augmentation method where \textit{hard} examples are added to the training set. Therefore, besides just having models that are trained on clean adversarial examples (i.e. \quotes{natural training}) as our baseline, we also compare our results to models trained using more conventional data augmentation methods. We use SSMBA \cite{ng-etal-2020-ssmba} and backtranslation\footnote{For backtranslation, we use English-to-German model and German-to-English model trained by \citet{ng2019facebook}.} \cite{uda} methods as our baselines as both have reported strong performance on text classification tasks. We use these methods to generate approximately the same number of new training examples as adversarial training.

\subsection{Results on Adversarial Robustness}
To evaluate models' robustness to adversarial attacks, we attempt to generate adversarial examples from 1000 randomly sampled clean examples from the test set and measure the attack success rate.
$$\text{attack success rate} = \frac{\text{\# of successful attacks}}{\text{\# of total attacks}}$$ 

Table \ref{table:attack-emb-robustness} shows the attack success rates of \fasttextfooler attack and \fastbae attack against models that have been trained using \method\!, \methodV\!, and other baseline methods. Note that the overall attack success rates appear fairly low because we applied strict constraints to improve the quality of the adversarial examples (as recommend by \citet{morris2020reevaluating}). Still, we can see that for both attacks, adversarial training using the same attack can decrease the attack success rate by up to 70\%. What is surprising is that training the model using a different attack also led to a decrease in the attack success rate. From Table \ref{table:attack-emb-robustness}, we can see that adversarial training using the \fastbae attack lowers the attack success rate of \fasttextfooler attack while training with \fasttextfooler lowers the attack success rate of \fastbae attack.

To further measure how adversarial training using \method\! can improve model's robustness, we evaluated accuracy of BERT-\method on 1000 adversarial examples that have successfully fooled BERT-Natural models. Table \ref{table:attack-accuracy} shows that adversarial training using \method greatly improves model's performance against adversarial examples from baseline 0\% to over 70\% for IMDB and Yelp datasets.

Another surprising observation is that training with data augmentations methods like SSMBA and backtranslation can lead to improvements in robustness against both adversarial attacks. However, in case of smaller datasets such as MR, data augmentation can also hurt robustness.

When we compare the attack success rates between BERT and RoBERTa models in Table \ref{table:attack-emb-robustness}, we also see an interesting pattern. BERT models, regardless of the training method, tend to be more vulnerable to \method attack than RoBERTa models. At the same time, RoBERTa models tend to be more vulnerable to \fastbae attack than BERT models.

Lastly, we use attacks proposed from literature to evaluate the models' adversarial robustness. Table \ref{table:robustness-literature} shows the attack success rate of TextFooler \cite{Jin2019TextFooler}, BAE \cite{garg2020bae}, PWWS \cite{pwws-ren-etal-2019-generating}, and PSO \cite{pso-zang-etal-2020-word}.\footnote{These attacks were implemented using the \TextAttack library \cite{Morris2020TextAttackAF}.} Across four datasets and two models, we can see that both \fasttextfooler and \fastbae lower the attack success rate against all four attacks in all but five cases. The results for PWWS and PSO are especially surprising since both use different transformations - WordNet \cite{wordnet} and HowNet \cite{HowNet-2010} - when carrying out the attacks.

\begin{table*}[]
\centering
\scalebox{0.8}{
\begin{tabular}{clcrcrcrcr}
\toprule[0.25ex]
\multirow{2}{*}{ Attack } & \multirow{2}{*}{ Model } & \multicolumn{2}{c}{IMDB} & \multicolumn{2}{c}{MR} & \multicolumn{2}{c}{Yelp} & \multicolumn{2}{c}{SNLI} \\
\cmidrule(lr){3-4} \cmidrule(lr){5-6} \cmidrule(lr){7-8} \cmidrule(lr){9-10}
& & A.S.\% & $\Delta\%$ & A.S.\% & $\Delta\%$ & A.S.\% & $\Delta\%$ & A.S.\% & $\Delta\%$ \\
\cmidrule(lr){1-10}
\multirow{10}{*}{ \fasttextfooler } & BERT-Natural & 42.9 &  & 20.9 &  & 25.4 &  & 53.3 &  \\
& BERT-\fasttextfooler & \textbf{12.7} & \textbf{-70.4} & \textbf{13.2} & \textbf{-36.8} & \textbf{11.5} & \textbf{-54.7} & \textbf{15.6} & \textbf{-70.7} \\
& BERT-\fastbae & 34.5 & -19.6 & 18.9 & -9.6 & 21.0 & -17.3 & 47.2 & -11.4 \\
& BERT-SSMBA & 29.5 & -31.2 & 21.1 & 1.0 & 23.3 & -8.3 & 51.0 & -4.3 \\
& BERT-BackTranslation & 33.1 & -22.8 & 19.2 & -8.1 & 24.0 & -5.5 & 48.3 & -9.4 \\
\cmidrule(lr){2-10}
& RoBERTa-Natural & 34.3 &  & 18.6 &  & 19.9 &  & 48.4 & \\
& RoBERTa-\fasttextfooler & \textbf{12.4} & \textbf{-63.8} & \textbf{12.1} & \textbf{-34.9} & \textbf{7.6} & \textbf{-61.8} & \textbf{8.3} & \textbf{-82.9}\\
& RoBERTa-\fastbae & 19.5 & -43.1 & 17.1 & -8.1 & 13.0 & -34.7 & 40.3 & -16.7\\
& RoBERTa-SSMBA & 24.0 & -30.0 & 21.8 & 17.2 & 19.3 & -3.0 & 48.7 & 0.6 \\
& RoBERTa-BackTranslation & 28.9 & -15.7 & 18.3 & -1.6 & 16.1 & -19.1 & 48.6 & 0.4 \\
\cmidrule(lr){1-10}
\multirow{10}{*}{ \fastbae } & BERT-Natural & 76.6 &  & 37.7 &  & 47.1 &  & 77.9 &  \\
& BERT-\fasttextfooler & 61.7 & -19.5 & 33.2 & -11.9 & 42.5 & -9.8 & 76.7 & -1.5 \\
& BERT-\fastbae & \textbf{48.3} & \textbf{-36.9} & \textbf{24.7} & \textbf{-34.5} & \textbf{27.9} & \textbf{-40.8} & \textbf{37.1} & \textbf{-52.4} \\
& BERT-SSMBA & 59.6 & -22.2 & 36.2 & -4.0 & 44.8 & -4.9 & 76.9 & -1.3 \\
& BERT-BackTranslation & 68.8 & -10.2 & 36.3 & -3.7 & 46.8 & -0.6 & 77.3 & -0.8 \\
\cmidrule(lr){2-10}
& RoBERTa-Natural & 81.5 &  & 40.9 &  & 53.2 &  & 78.6 &  \\
& RoBERTa-\fasttextfooler & 69.8 & -14.4 & 38.4 & -6.1 & 45.2 & -15.0 & 76.5 & -2.7 \\
& RoBERTa-\fastbae & \textbf{37.0} & \textbf{-54.6} & \textbf{28.5} & \textbf{-30.3} & \textbf{25.8} & \textbf{-51.5} & \textbf{35.2} & \textbf{-55.2} \\
& RoBERTa-SSMBA & 57.0 & -30.1 & 43.1 & 5.4 & 47.8 & -10.2 & 78.3 & -0.4 \\
& RoBERTa-BackTranslation & 74.3 & -8.8 & 41.1 & 0.5 & 43.8 & -17.7 & 79.1 & 0.6 \\
\bottomrule[0.25ex]
\end{tabular}
}
\caption{Attack success rate of \fasttextfooler and \fastbae attacks. A.S.\% represents the attack success rates and $\Delta\%$ column represents the percent change between the attack success rate of natural training and the different training methods. }
\label{table:attack-emb-robustness}
\end{table*}

\begin{table*}[]
\centering
\scalebox{0.85}{
\begin{tabular}{clcrcrcrcr}
\toprule[0.25ex]
\multirow{2}{*}{ Attack } & \multirow{2}{*}{ Model } & \multicolumn{2}{c}{IMDB} & \multicolumn{2}{c}{MR} & \multicolumn{2}{c}{Yelp} & \multicolumn{2}{c}{SNLI} \\
\cmidrule(lr){3-4} \cmidrule(lr){5-6} \cmidrule(lr){7-8} \cmidrule(lr){9-10}
& & A.S.\% & $\Delta \%$ & A.S.\% & $\Delta \%$ & A.S.\% & $\Delta \%$ & A.S.\% & $\Delta \%$ \\
\cmidrule(lr){1-10}
\multirow{6}{*}{ TextFooler } & BERT-Natural & 85.0 &  & 91.6 &  & \textbf{55.9} &  & 97.5 &  \\
& BERT-\method & \textbf{66.0} & \textbf{-22.4} & 90.6 & -1.1 & 57.9 & 3.6 & \textbf{92.2} & \textbf{-5.4} \\
& BERT-\methodV & 88.2 & 3.8 & \textbf{89.0} & \textbf{-2.8} & 67.7 & 21.1 & 94.1 & -3.5 \\
\cmidrule(lr){2-10}
& RoBERTa-Natural & 95.2 &  & 94.4 &  & 74.5 &  & 96.7 &  \\
& RoBERTa-\method & \textbf{82.4} & \textbf{-13.4} & \textbf{91.0} & \textbf{-3.6} & \textbf{68.7} & \textbf{-7.8} & 91.4 & -5.5 \\
& RoBERTa-\methodV & 72.9 & -23.4 & 88.6 & -6.1 & 71.7 & -3.8 & \textbf{90.8} & \textbf{-6.1} \\
\cmidrule(lr){1-10}
\multirow{6}{*}{ BAE } & BERT-Natural & 60.5 &  & 52.6 &  & 37.8 &  & 76.7 &  \\
& BERT-\method & \textbf{46.7} & \textbf{-22.8} & 51.5 & -2.1 & 34.4 & -9.0 & 75.9 & -1.0 \\
& BERT-\methodV & 52.4 & -13.4 & \textbf{43.8} & \textbf{-16.7} & \textbf{31.3} & \textbf{-17.2} & \textbf{60.9} & \textbf{-20.6} \\
\cmidrule(lr){2-10}
& RoBERTa-Natural & 65.5 &  & 56.4 &  & 44.4 &  & 75.6 &  \\
& RoBERTa-\method & 56.8 & -13.3 & 54.7 & -3.0 & 38.0 & -14.4 & 76.0 & 0.5 \\
& RoBERTa-\methodV & \textbf{42.3} & \textbf{-35.4} & \textbf{48.3} & \textbf{-14.4} & \textbf{28.7} & \textbf{-35.4} & \textbf{61.2} & \textbf{-19.0} \\
\cmidrule(lr){1-10}
\multirow{6}{*}{ PWWS } & BERT-Natural & 87.5 &  & 82.1 &  & 67.9 &  & 98.5 &  \\
& BERT-\method & \textbf{70.9} & \textbf{-19.0} & \textbf{80.4} & \textbf{-2.1} & \textbf{65.4} & \textbf{-3.7} & \textbf{97.5} & \textbf{-1.0} \\
& BERT-\methodV & 87.1 & -0.5 & 81.3 & -1.0 & 72.2 & 6.3 & \textbf{97.5} & \textbf{-1.0} \\
\cmidrule(lr){2-10}
& RoBERTa-Natural & 96.6 &  & 83.8 &  & 77.9 &  & 98.2 &  \\
& RoBERTa-\method & 84.4 & -12.6 & 81.9 & -2.3 & 73.1 & -6.2 & 97.1 & -1.1 \\
& RoBERTa-\methodV & \textbf{73.5} & \textbf{-23.9} & \textbf{79.8} & \textbf{-4.8} & \textbf{70.7} & \textbf{-9.2} & \textbf{96.5} & \textbf{-1.7} \\
\cmidrule(lr){1-10}
\multirow{6}{*}{ PSO } & BERT-Natural & 43.8 &  & 81.6 & -& 40.3 &  & 92.1 &  \\
& BERT-\method & \textbf{16.5} & \textbf{-62.3} & \textbf{73.2} & \textbf{-10.3} & \textbf{26.4} & \textbf{-34.5} & \textbf{89.1} & \textbf{-3.3} \\
& BERT-\methodV & 29.9 & -31.7 & 75.4 & -7.6 & 34.4 & -14.6 & 89.7 & -2.6 \\
\cmidrule(lr){2-10}
& RoBERTa-Natural & 34.8 &  & 88.0 &  & 35.7 &  & 90.6 &  \\
& RoBERTa-\method & \textbf{12.9} & \textbf{-62.9} & 81.6 & -7.3 & 21.6 & -39.5 & 85.3 & -5.8 \\
& RoBERTa-\methodV & 13.1 & -62.4 & \textbf{77.5} & \textbf{-11.9} & \textbf{20.3} & \textbf{-43.1} & \textbf{84.8} & \textbf{-6.4} \\
\bottomrule[0.25ex]
\end{tabular}
}
\caption{Attack success rate of attacks from literature, including original TextFooler \cite{Jin2019TextFooler}, BAE \cite{garg2020bae}, PWWS \cite{pwws-ren-etal-2019-generating}, and PSO \cite{pso-zang-etal-2020-word}.  A.S.\% represents the attack success rates and $\Delta\%$ column represents the percent change between natural training and the different training methods.}
\label{table:robustness-literature}
\end{table*}

\begin{table*}[]
\centering
\scalebox{0.85}{
\begin{tabular}{ccrrrrrr}
\toprule[0.25ex]
Dataset & Model & A2T & A2T-MLM & TextFooler & BAE & PSO & PWWS \\
\cmidrule(lr){1-8}
IMDB & BERT-\method & 94.60 & 75.87 & 93.96 & 78.42 & 87.00 & 89.14 \\
MR & BERT-\method & 65.90 & 53.25 & 57.56 & 31.96 & 36.57 & 47.76 \\
Yelp & BERT-\method & 92.00 & 77.78 & 87.17 & 73.88 & 72.44 & 83.21 \\
SNLI & BERT-\method & 56.43 & 47.10 & 41.99 & 42.71 & 41.67 & 39.84 \\
\bottomrule[0.25ex]
\end{tabular}
}
\caption{Accuracy of BERT-\method on 1000 adversarial examples that have successfully fooled BERT-Natural.}
\label{table:attack-accuracy}
\end{table*}

\begin{table*}[]
\centering
\scalebox{0.80}{
\begin{tabular}{lcccccccc}
\toprule[0.25ex]
\multirow{2}{*}{ Model } & \multicolumn{2}{c}{IMDB} & \multicolumn{2}{c}{MR} & \multicolumn{2}{c}{Yelp} & \multicolumn{2}{c}{SNLI} \\
\cmidrule(lr){2-3} \cmidrule(lr){4-5} \cmidrule(lr){6-7} \cmidrule(lr){8-9}
&\begin{tabular}[c]{@{}c@{}}Standard\\ Accuracy\end{tabular} & \begin{tabular}[c]{@{}c@{}}Yelp\\ Accuracy\end{tabular}  &
\begin{tabular}[c]{@{}c@{}}Standard\\ Accuracy\end{tabular} &
\begin{tabular}[c]{@{}c@{}}Yelp\\ Accuracy\end{tabular}  &
\begin{tabular}[c]{@{}c@{}}Standard\\ Accuracy\end{tabular} &
\begin{tabular}[c]{@{}c@{}}IMDB\\ Accuracy\end{tabular}  &
\begin{tabular}[c]{@{}c@{}}Standard\\ Accuracy\end{tabular} &
\begin{tabular}[c]{@{}c@{}}MNLI\\ Accuracy\end{tabular}\\
\cmidrule(lr){1-9}
BERT-Natural & 93.97 & 92.13 & 85.40 & \textbf{90.60} & 96.34 & 88.31 & 90.29 & 73.34 \\
BERT-\fasttextfooler & \textbf{94.49} & \textbf{92.50 }& 85.61 & 88.45 & \textbf{96.68} & \textbf{89.24} & 90.16 & \textbf{73.79} \\
BERT-\fastbae & 93.05 & 90.67 & 83.80 & 85.32 & 95.85 & 85.01 & 87.87 & 70.93 \\
BERT-SSMBA & 93.94 & 91.59 & 85.33 & 89.49 & 96.28 & 88.54 & 90.23 & 73.27 \\
BERT-BackTranslation & 93.97 & 91.73 & \textbf{85.65} & 89.46 & 96.46 & 88.77 & \textbf{90.57} & 72.82 \\
\cmidrule(lr){1-9}
RoBERTa-Natural & 95.26 & 94.09 & 87.52 & 93.42 & 97.26 & \textbf{91.94} & \textbf{91.56} & 77.66 \\
RoBERTa-\fasttextfooler & \textbf{95.57} & 94.41 & \textbf{88.03} & 93.45 & \textbf{97.45} & 91.86 & 91.16 & \textbf{77.88} \\
RoBERTa-\fastbae & 94.71 & \textbf{94.48} & 86.49 & 92.93 & 96.84 & 90.44 & 88.56 & 74.82 \\
RoBERTa-SSMBA & 95.25 & 94.11 & 86.46 & 93.03 & 97.16 & 91.90 & 91.38 & 77.02 \\
RoBERTa-BackTranslation & 95.31 & 93.84 & 87.78 & \textbf{93.77} & 97.25 & 91.76 & 90.79 & 76.70 \\ 
\bottomrule[0.25ex]
\end{tabular}
}
\caption{Accuracy on in-domain and out-of-domain datasets. We can see that adversarial training can helps model outperform both naturally trained models and models trained using data augmentation methods.}
\label{table:accuracy}
\end{table*}

\begin{table}[ht]
\centering
\scalebox{0.75}{
\begin{tabular}{lrrr}
\toprule[0.25ex]
Model & IMDB & MR & Yelp \\
\cmidrule(lr){1-4}
BERT-Natural & 7.78 & 33.43 & 12.78  \\
BERT-\fasttextfooler & \textbf{10.74} & \textbf{34.25} & \textbf{13.18} \\
BERT-\fastbae & 9.12 & 32.17 & 11.14 \\
BERT-SSMBA & 7.21 & 32.21 & 10.94\\
BERT-BackTranslation & 6.02 & 0.39 & 11.10 \\
\cmidrule(lr){1-4}
RoBERTa-Natural & 0.35 & 0.01 & -1.09 \\
RoBERTa-\fasttextfooler & \textbf{0.43} & \textbf{0.45} & -1.01 \\
RoBERTa-\fastbae & 0.09 & -0.12 & -1.13 \\
RoBERTa-SSMBA & 0.26 & 0.05 & \textbf{-0.43 }\\
RoBERTa-BackTranslation & -0.04 & 0.05 & -1.06 \\
\bottomrule[0.25ex]
\end{tabular} 
}
\caption{AOPC scores of the LIME explanations for each model. Higher AOPC scores indicates that the model is more interpretable.}
\label{table:aopc-scores}
\end{table}

\subsection{Results on Generalization}
To evaluate how adversarial training affects the model's generalization ability, we evaluate its accuracy on the original test set (i.e. standard accuracy) and on an out-of-domain dataset (e.g. Yelp dataset for model trained on IMDB dataset). In Table \ref{table:accuracy}, we can see that in all but two cases, adversarial training using \fasttextfooler attack beats natural training in terms of standard accuracy. In the two cases (SNLI) where natural training beats \method\!, we can see that \method still outperforms natural training in cross-domain accuracy. Overall, in six out of eight cases, \fasttextfooler  improves cross-domain accuracy. On the other hand, adversarial training with \fastbae attack tends to hurt both standard accuracy and cross-domain accuracy. This confirms the observations reported by \citet{clare2021} and suggests that using a masked language model to generate adversarial examples can lead to a trade-off between robustness and generalization. We do not see similar trade-off with \fasttextfooler\!.

\subsection{Results on Interpretability}
We use LIME \cite{lime} to generate local explanations for our models. For each example, LIME approximates the local decision boundary by fitting a linear model over the samples obtained by perturbing the example. To measure the faithfulness of the local explanations obtained using LIME, we measure the area over perturbation curve (AOPC) \cite{samek2017, nguyen-2018-comparing, chen-ji-2020-learning} which is defined as:
\begin{equation}
    AOPC = \frac{1}{K+1} \sum^K_{k=1} \frac{1}{N} \sum^N_{i=1} f(\x^{(i)}_{(0)}) - f(\x^{(i)}_{(k)})
\end{equation}
where $\x^{(i)}_{(0)}$ represents example $\x^{(i)}$ with none of the words removed and  $\x^{(i)}_{(k)}$ represents example $\x^{(i)}$ with the top-$k$ most important words removed. $f(\x)$ here represents the model's confidence on the target label $\y^{(i)}$.  Intuitively, AOPC measures on average how the model's confidence on the target label changes when we delete the top-$k$ most important words determined using LIME.

For each dataset, we randomly pick 1000 examples from the test set for evaluation. When running LIME to obtain explanations, we generate $1000$ perturbed samples for each instance. We set $K=10$ for the AOPC metric. Table \ref{table:aopc-scores} shows that across three sentiment classification datasets, BERT model trained using \fasttextfooler attack achieves higher AOPC than natural training. For RoBERTa models, the same observation holds (although by smaller margins). Overall, we see that the AOPC scores for RoBERTa models are far lower than those for BERT models, suggesting that RoBERTa might be less interpretable than BERT.

\begin{figure}[]
\centering
\begin{minipage}[]{0.49\hsize}
\centering
\includegraphics[width=\hsize]{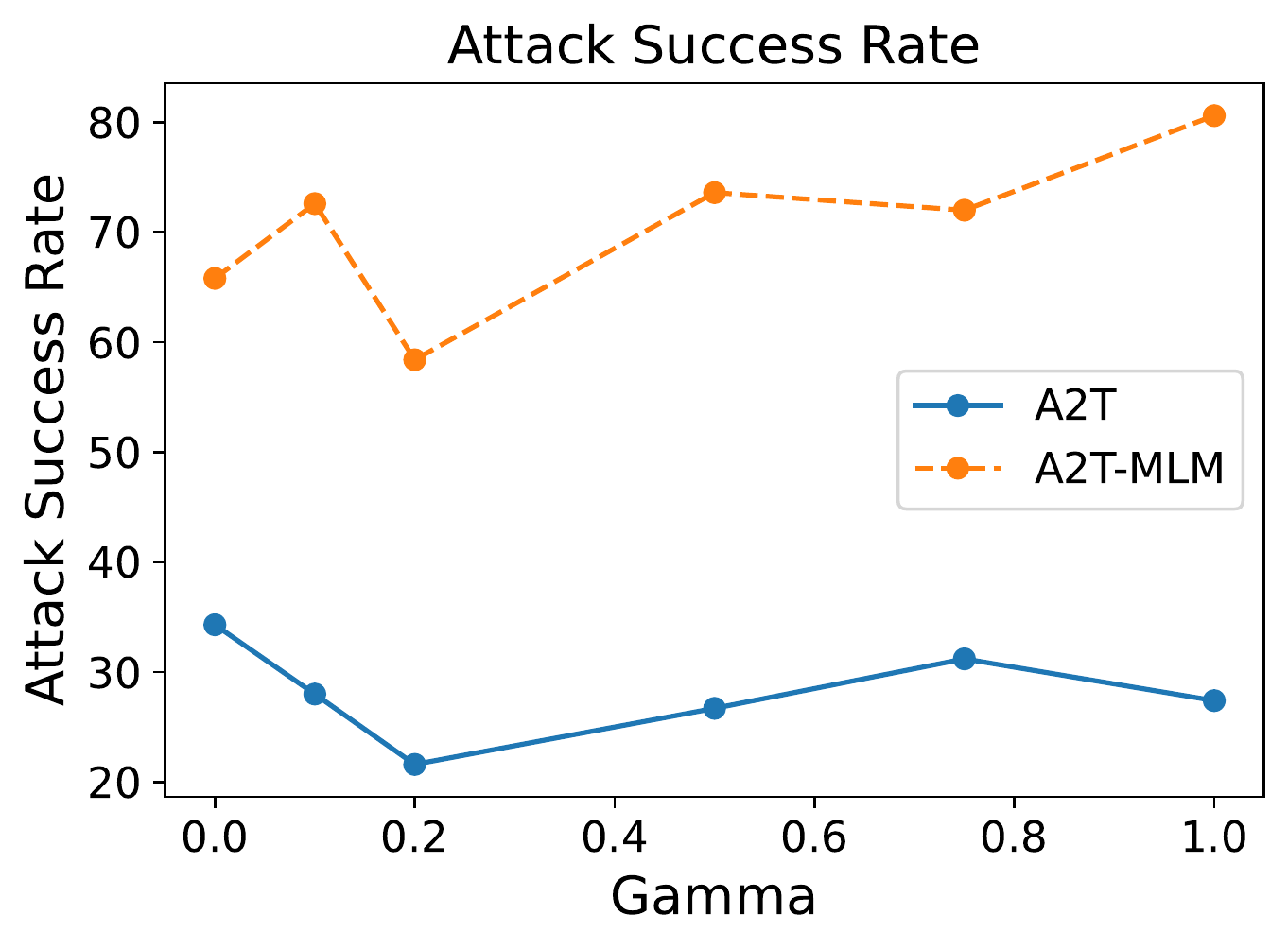}
(a)
\end{minipage}
\begin{minipage}[]{0.49\hsize}
\centering
\includegraphics[width=\hsize]{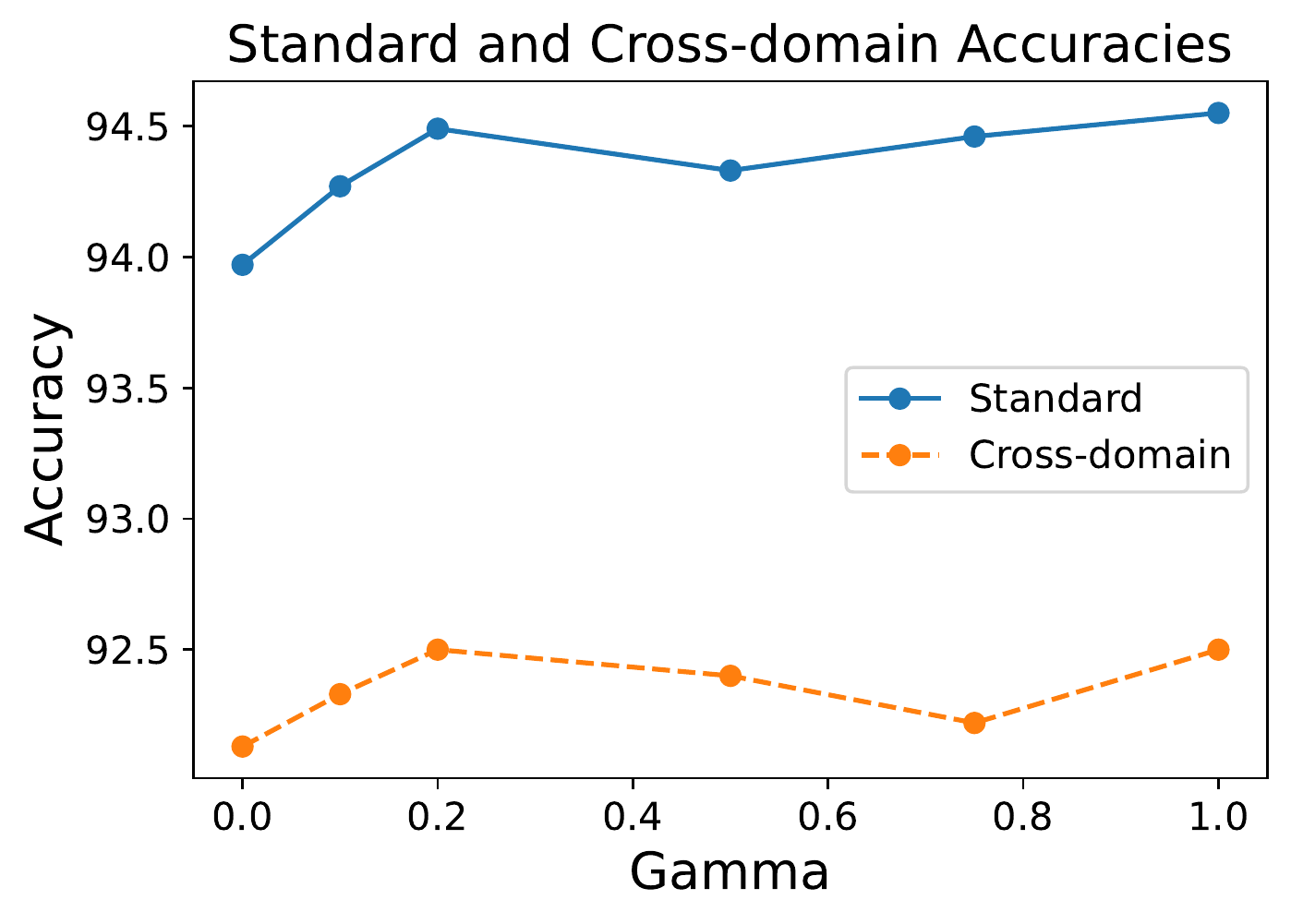}
(b)
\end{minipage}
\caption{Left: Attack success rates of \method and \methodV attacks on BERT-\method model trained on IMDB dataset. Right: Standard accuracies and cross-domain accuracies (on Yelp) for the same model.\label{gamma-figure}}
\vspace{-5mm}
\end{figure}

\subsection{ Analysis}
\paragraph{\fasttextfooler vs \fastbae attack}
We can see that model trained using \fasttextfooler attack outperforms the model trained using \fastbae attack in standard accuracy and cross-domain accuracy in all but one case. This suggests that using counter-fitted embeddings can generate higher quality adversarial examples than masked language models. Since masked language models are only trained to predict words that are statistically most likely to appear, it is likely that it will propose words that do change the semantics of the text entirely; this can lead to false positive adversarial examples. 

We also hypothesize that \methodV\!'s tendency to generate false positive adversarial examples is the reason why RoBERTa appears to be more vulnerable to \methodV than BERT models. Since RoBERTa models tend to have better generalization capability than BERT models, RoBERTa models are more likely to predict the correct labels for false positive adversarial examples that \methodV can generate (whereas BERT models can predict the wrong labels for false positive adversarial examples and appear falsely robust).

\paragraph{Effect of Gamma}
\label{gamma-study}

Recall $\gamma$, which is the desired percentage of adversarial examples to generate in every epoch. To study how it affects the results of adversarial training, we train BERT model on IMDB dataset with \fasttextfooler method and $\gamma$ value ranging from  $0$ (no adversarial training) to $1.0$. Figure \ref{gamma-figure} shows the attack success rates, standard accuracies, and cross-domain accuracies.

We can see from Figure \ref{gamma-figure} (a) that higher $\gamma$ does not necessarily mean that our trained model is more robust, with $\gamma=0.2$ producing model that is more robust than others. Overall, we can see that adversarial training is insensitive to the specific choice of $\gamma$ and a smaller $\gamma$ can be used for faster training.

\paragraph{Effect of Adversarial Training on Sentence Embedding}
\begin{figure*}[]
\centering
\includegraphics[width=\hsize]{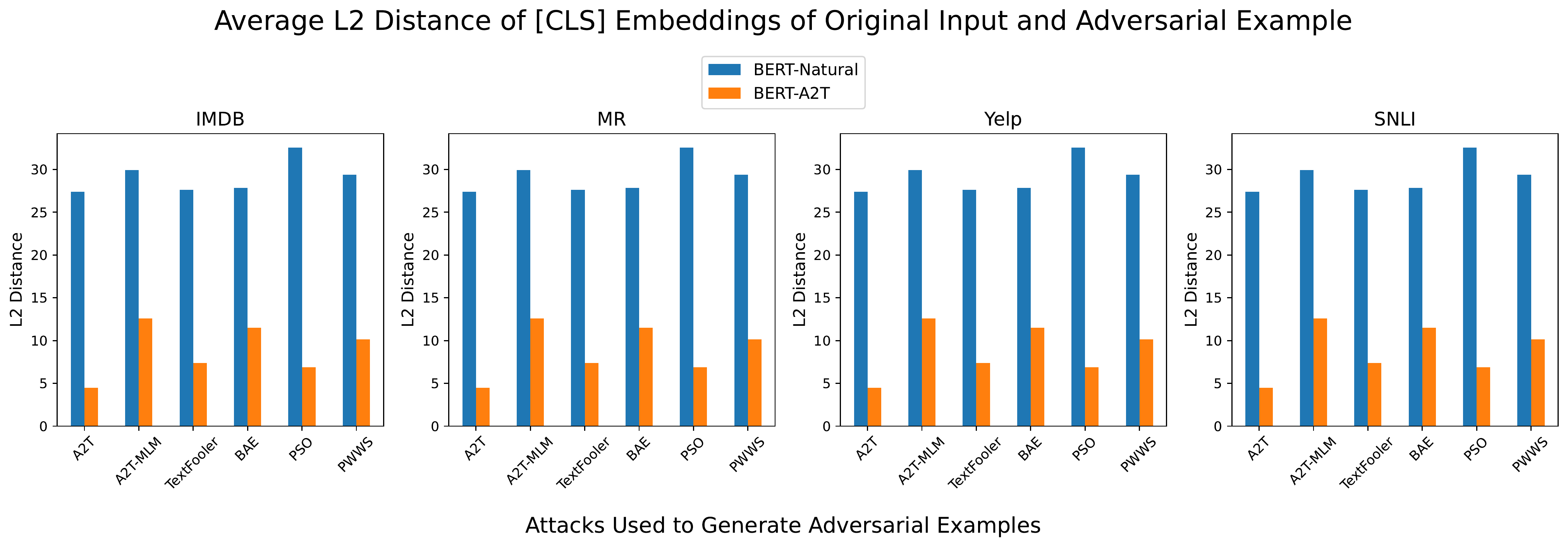}
\caption{BERT-\method decreases the $\ell_2$ distance between \texttt{[CLS]} embeddings of original $\x$ and adversarial $\x'$. \label{fig:cls-figure}}
\end{figure*}

In BERT \cite{devlin2018BERT}, the output for \texttt{[CLS]} token represents the sentence-level embedding that is used by the final classification layer. For BERT-Natural and BERT-\method\!, we measured the $\ell_2$ distance between the \texttt{[CLS]} embeddings of original text $\x$ and its corresponding adversarial example $\x'$ using six different attacks\footnote{The adversarial examples used were the same examples used to generate results for Table \ref{table:attack-accuracy}}. We noticed that across all cases, adversarial training decreases the average $\ell_2$ distance between $\x$ and $\x'$, as shown by Figure \ref{fig:cls-figure}. This suggests that adversarial training improves the robustness of models by encouraging the model to learn a closer mapping of $\x$ and $\x'$.

\section{Conclusion}
In this paper, we have presented a practical \base process called \method that uses a new adversarial attack designed to generate adversarial examples quickly. We demonstrated that using \method allows us to improve model's robustness against several different types of adversarial attacks that have been proposed from literature. Also, we have shown that models trained using \method can achieve better standard accuracy and/or cross-domain accuracy than baseline models.

\bibliography{references}
\bibliographystyle{acl_natbib}

\appendix
\clearpage
\section{Appendix}

\subsection{\method and \methodV Attacks}
Here, we give more details about \method and \methodV. We will use the framework introduced by \cite{Morris2020TextAttackAF} to break down adversarial attacks into the following four components: (1) goal function, (2) transformation, (3) a set of constraints, (4) search method.

\paragraph{Goal Function}
In this work, we perform untargeted attack since they are generally easier than targeted attack. We aim to maximize the following as our goal function:
\begin{equation}
    1 - P(y | \x; \theta) 
\end{equation}
where $P(y | \x; \theta)$ means the model's confidence of label $y$ given input $\x$ and parameters $\theta$.

\paragraph{Transformation}
\begin{enumerate}
    \item \method Counter-fitted word embedding \cite{Mrksic2016CounterfittingWV}
    \item \methodV: BERT masked language model \cite{devlin2018BERT}
\end{enumerate}

For both methods, we select the top 20 words proposed as replacements. This helps us narrow down our replacements to the best ones and save time from considering less desirable replacements.

\paragraph{Constraints}
We use the following constraints for both attacks:
\begin{itemize}
    \item \textbf{Part-of-speech Consistency}:
        To preserve fluency, we require that the two words being swapped have the same part-of-speech. This is determined by a part-of-speech tagger provided by Flair \cite{flair-akbik2018coling}, an open-source NLP library.
    \item \textbf{DistilBERT Semantic Textual Similarity (STS)} \cite{distil-bert}:
        We require that cosine similarity between the sentence encodings of original text $\x$ and perturbed text $\x'$ meet minimum threshold value of $0.9$. We use fine-tuned DistilBERT model provided by \citet{distilbert-sts}.
    \item \textbf{Max modification rate}: 
        We allow only 10\% of the words to be replaced. This limits us from modifying the text too much and causing the semantics of the text to change.
\end{itemize}

Also, for \fasttextfooler attack, we require that the word embeddings between original text $\x$ and perturbed text $\x'$ have minimum cosine similarity of $0.8$. 

The threshold values for word embedding similarity and sentence encoding similarity were set based on the recommendations by \citet{morris2020reevaluating}, which noted that high threshold values encourages strong semantic similarity between the original text and the perturbed text.

\paragraph{Search Method}
Search method is responsible for iteratively perturbing the original text $\x$ until we discover an adversarial example $\x_{adv}$ that causes the model to mispredict. Algorithm \ref{alg:search-method} shows \method's search algorithm. If the search method fails to find an adversarial example by the time its search is over, it has failed to generate one. It can also exit preemptively if it has reached maximum number of queries to the victim model. Such limit is called query budget.

\begin{algorithm}[t]
\caption{\method's Search Method: Gradient-based Word Importance Ranking}
\begin{algorithmic}[1]
    \REQUIRE Original text $\x=(x_1, x_2, ... x_n)$. Transformation module $T(\x, i)$ that perturbs $\x$ by replacing $x_i$.
    \ENSURE Adversarial text $\x_{adv}$ if found
    \STATE Calculate $I(x_i)$ for all words $x_i$ by making one forward and backward pass.
    \STATE $R \leftarrow $ ranking $r_1,\dots,r_n$ of words $x_1,\dots,x_n$ by descending importance
    \STATE $\x^*\leftarrow \x$
    \FOR {$i = r_1, r_2,\dots, r_n$ in $R$}
        \STATE $X_{cand} \leftarrow T(x^*, i)$
        \IF{$X_{cand} \neq \emptyset$}
            \STATE $\x^* \leftarrow \argmax_{\x'\in X_{cand}}{1 - P(y | \x; \theta)} $
            \IF{$\x^*$ fools the model}
                \RETURN{$\x^*$ as $\x_{adv}$}
            \ENDIF
        \ENDIF
    \ENDFOR
\end{algorithmic}
\label{alg:search-method}
\end{algorithm}

During training, we limit the search method to making only 200 queries to the victim model for faster generation of adversarial examples. For evaluation using \method and \methodV, we increase the query budget to 2000 queries for a more extensive search. For other attacks such as TextFooler \cite{Jin2019TextFooler}, BAE \cite{garg2020bae}, PWWS \cite{pwws-ren-etal-2019-generating}, and PSO \cite{pso-zang-etal-2020-word}, the query budget is set to 5000.

\end{document}